\PassOptionsToPackage{unicode}{hyperref}
\PassOptionsToPackage{hyphens}{url}
\PassOptionsToPackage{dvipsnames,svgnames,x11names}{xcolor}
\documentclass[
  10pt,
]{article}
\usepackage{xcolor}
\usepackage[letterpaper,margin=1in]{geometry}
\usepackage{amsmath,amssymb}
\usepackage[T1]{fontenc}
\usepackage[utf8]{inputenc}
\usepackage{textcomp}
\usepackage{tgtermes}
\IfFileExists{upquote.sty}{\usepackage{upquote}}{}
\IfFileExists{microtype.sty}{
  \usepackage[]{microtype}
  \UseMicrotypeSet[protrusion]{basicmath} 
}{}
\usepackage{longtable,booktabs,array}
\usepackage{calc} 
\usepackage{etoolbox}
\makeatletter
\patchcmd\longtable{\par}{\if@noskipsec\mbox{}\fi\par}{}{}
\makeatother
\IfFileExists{footnotehyper.sty}{\usepackage{footnotehyper}}{\usepackage{footnote}}
\makesavenoteenv{longtable}
\providecommand{\tightlist}{%
  \setlength{\itemsep}{0pt}\setlength{\parskip}{0pt}}

\usepackage{graphicx,booktabs,float,microtype,newunicodechar,xurl}
\providecommand{\bibsep}{\itemsep}

\usepackage[font=small,labelfont=bf,labelsep=period]{caption}
\makeatletter\def\fps@figure{htbp}\def\fps@table{htbp}\makeatother
\makeatletter\renewcommand\maketitle{\begin{center}{\LARGE\bfseries\@title\par}\vskip 1.2em{\large\@author\par}\vskip 0.6em{\itshape\@date\par}\end{center}\vskip 1em}\makeatother
\renewenvironment{abstract}{\begin{center}\bfseries\abstractname\end{center}\begin{quote}\small}{\end{quote}}
\newunicodechar{θ}{\ensuremath{\theta}}
\newunicodechar{π}{\ensuremath{\pi}}
\newunicodechar{ρ}{\ensuremath{\rho}}
\newunicodechar{η}{\ensuremath{\eta}}
\newunicodechar{α}{\ensuremath{\alpha}}
\newunicodechar{μ}{\ensuremath{\mu}}
\newunicodechar{ε}{\ensuremath{\varepsilon}}
\newunicodechar{τ}{\ensuremath{\tau}}
\newunicodechar{β}{\ensuremath{\beta}}
\newunicodechar{Σ}{\ensuremath{\Sigma}}
\newunicodechar{Δ}{\ensuremath{\Delta}}
\newunicodechar{≈}{\ensuremath{\approx}}
\newunicodechar{×}{\ensuremath{\times}}
\newunicodechar{·}{\ensuremath{\cdot}}
\newunicodechar{∇}{\ensuremath{\nabla}}
\newunicodechar{∼}{\ensuremath{\sim}}
\newunicodechar{√}{\ensuremath{\surd}}
\newunicodechar{≤}{\ensuremath{\le}}
\newunicodechar{≥}{\ensuremath{\ge}}
\newunicodechar{∈}{\ensuremath{\in}}
\newunicodechar{ℝ}{\ensuremath{\mathbb{R}}}
\newunicodechar{∞}{\ensuremath{\infty}}
\newunicodechar{⪯}{\ensuremath{\preceq}}
\newunicodechar{→}{\ensuremath{\to}}
\newunicodechar{±}{\ensuremath{\pm}}
\newunicodechar{⌊}{\ensuremath{\lfloor}}
\newunicodechar{⌋}{\ensuremath{\rfloor}}
\newunicodechar{…}{\ensuremath{\dots}}
\newunicodechar{−}{\ensuremath{-}}
\newunicodechar{⁰}{\ensuremath{^{0}}}
\newunicodechar{¹}{\ensuremath{^{1}}}
\newunicodechar{²}{\ensuremath{^{2}}}
\newunicodechar{³}{\ensuremath{^{3}}}
\newunicodechar{⁴}{\ensuremath{^{4}}}
\newunicodechar{⁵}{\ensuremath{^{5}}}
\newunicodechar{⁶}{\ensuremath{^{6}}}
\newunicodechar{⁷}{\ensuremath{^{7}}}
\newunicodechar{⁸}{\ensuremath{^{8}}}
\newunicodechar{⁹}{\ensuremath{^{9}}}
\newunicodechar{⁻}{\ensuremath{^{-}}}
\newunicodechar{₀}{\ensuremath{_{0}}}
\newunicodechar{§}{\S}
\usepackage{bookmark}
\IfFileExists{xurl.sty}{\usepackage{xurl}}{} 
\hypersetup{
  pdftitle={bf16 Rounding Silently Discards Most RL Policy Updates: A Hidden Training-Inference Mismatch and an Unbiased One-Line Fix},
  colorlinks=true,
  linkcolor={black},
  filecolor={Maroon},
  citecolor={Blue},
  urlcolor={blue},
  pdfcreator={LaTeX via pandoc}}

\title{bf16 Rounding Silently Discards Most RL Policy Updates: A Hidden Training-Inference Mismatch
and an Unbiased One-Line Fix}
\author{Andrew Gao\thanks{Corresponding author: \texttt{gaodrew@stanford.edu}} \quad William Gao\\[0.3em] Stanford University}
\date{Preprint}

\begin{document}
\maketitle
\begin{abstract}
Reinforcement learning (RL) post-training of language models keeps fp32 master weights in the
optimizer and serves rollouts from a bf16 copy produced by round-to-nearest conversion. At the
learning rates in common use, most per-step updates are smaller than half a bf16 spacing and the
conversion discards them: on Qwen3-1.7B only 0.4--0.5\% of served weights change per step, and the
policy that generates the training data lags the optimizer by 50--150 steps. Existing mismatch
corrections cannot see this lag because they compare two copies downstream of the same rounding. We
propose Fair Cast: export the master with an unbiased rounding rule (stochastic rounding or error
feedback) and compute the training loss on the served copy. The change is one line at the export
call, adds no hyperparameters and costs 22--38 ms per export of 1.7B parameters. Across 25 matched
seed-pairs on Qwen3-0.6B to 4B with AdamW and Muon, Fair Cast is ahead on 18, tied on 6 and behind
on 1, with a mean gain of 6.1 points; on zebra puzzles it reaches 0.57 ± 0.16 over four runs where
round-to-nearest collapses to 0.33 ± 0.10. It also makes the saved fp32 checkpoint agree with the
served model, and it exposes the implicit lag of round-to-nearest as a regularizer that can be
restored explicitly when a task needs it.
\end{abstract}

\begin{center}\textbf{Code:} \url{https://github.com/andrewgcodes/faircast}\end{center}

\begin{figure}[H]\centering\includegraphics[width=\linewidth,height=3.3in,keepaspectratio]{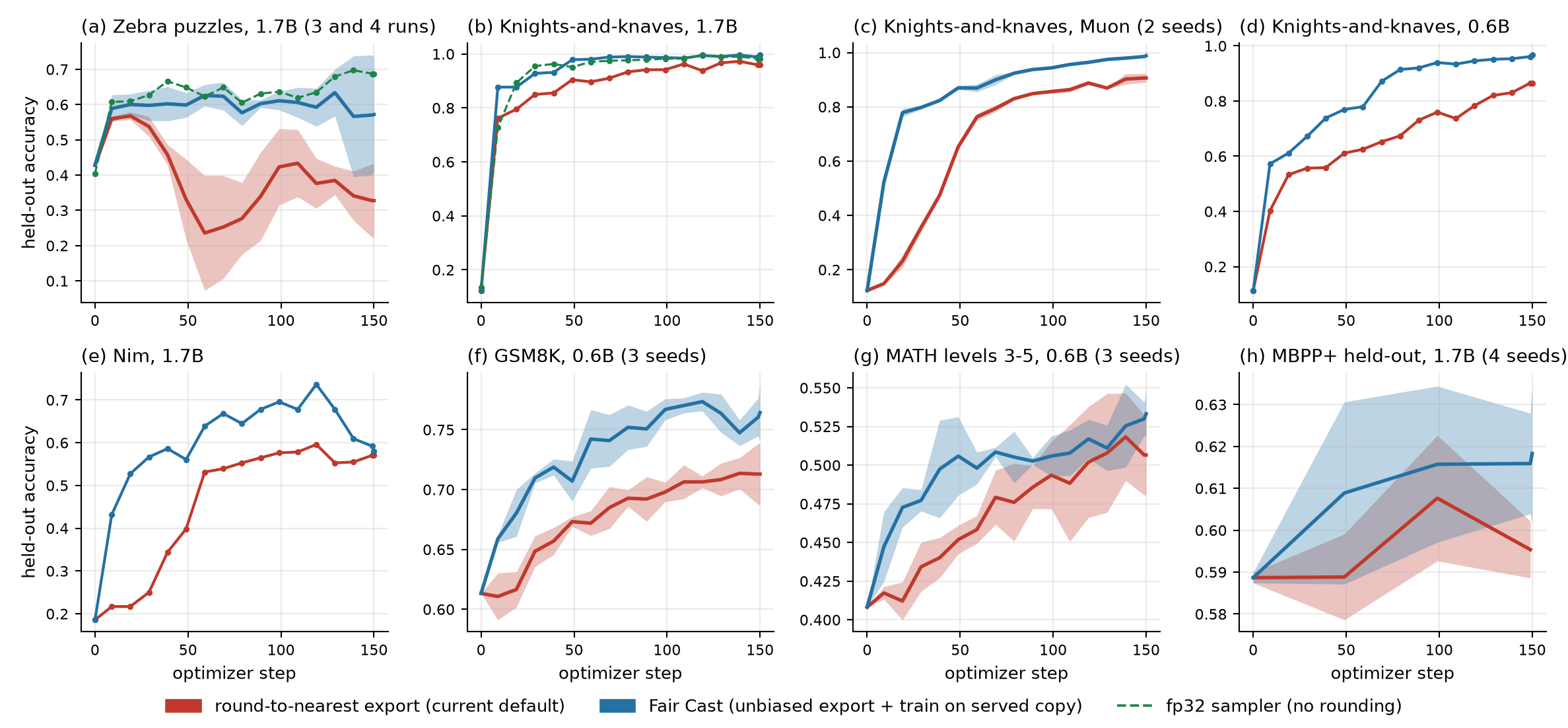}\caption{Fair Cast learns faster and more stably than round-to-nearest export across tasks, model sizes and optimizers, and tracks the fp32 sampler where one was run. Held-out accuracy of the served copy against optimizer step on logic (a–d), game (e), mathematics (f, g) and code (h) tasks; bands are ± 1 s.d. across runs (panel (a): three round-to-nearest and four Fair Cast runs; panel (h): scored on the 224 MBPP+ problems absent from the training split); the dashed line is a sampler run in fp32 with no rounding at all.}\label{fig:grid}\end{figure}

\section{Introduction}\label{introduction}

\begin{figure}[htbp]\centering\includegraphics[width=\linewidth]{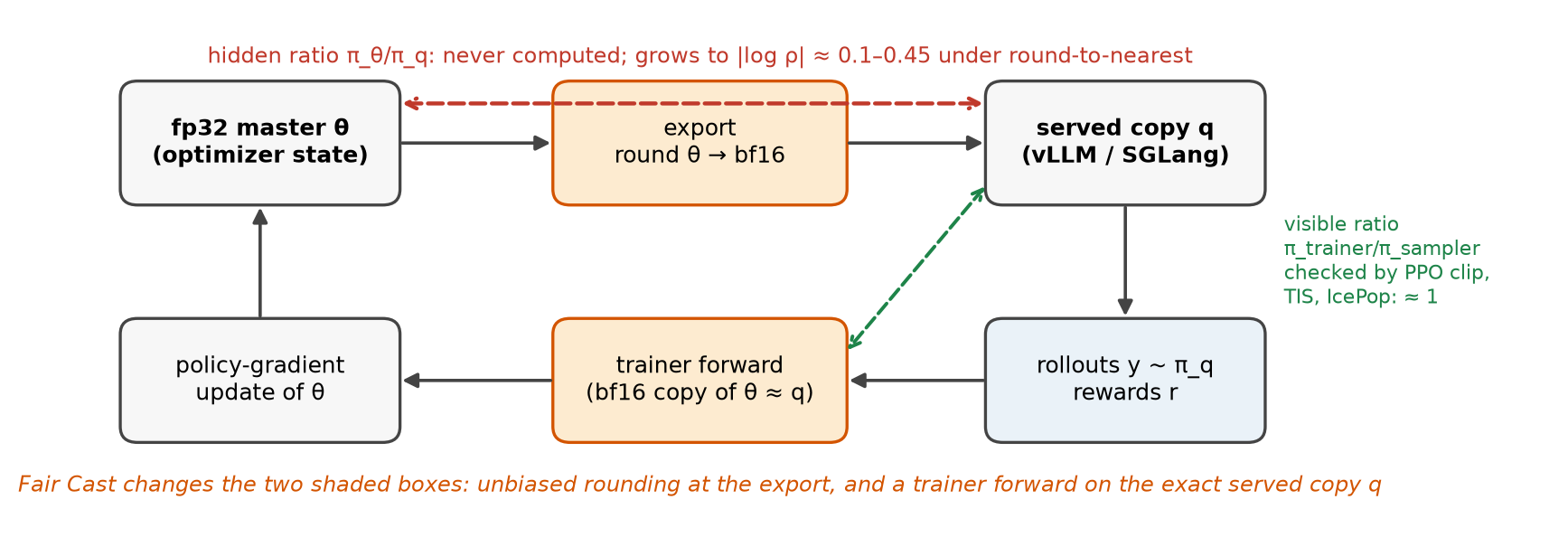}\caption{Where the two ratios live. The trainer's bf16 forward pass and the sampler both sit downstream of the export, so the ratio frameworks check (green) stays near one; the ratio between the fp32 master and the served copy (red) is never formed. Fair Cast changes the two shaded boxes.}\label{fig:pipeline}\end{figure}

Modern LLM reinforcement learning stacks separate the optimizer from the sampler. The optimizer
holds fp32 master weights θ; a dedicated inference engine (vLLM, SGLang) holds a bf16 copy q and
generates the rollouts. Every open-source framework we examined (verl, TRL, OpenRLHF, slime,
prime-rl, AReaL) produces q from θ by round-to-nearest conversion (Figure~\ref{fig:pipeline}). This paper is about
what that conversion does to learning, and how to fix it with one line of code.

\begin{figure}[htbp]\centering\includegraphics[width=\linewidth]{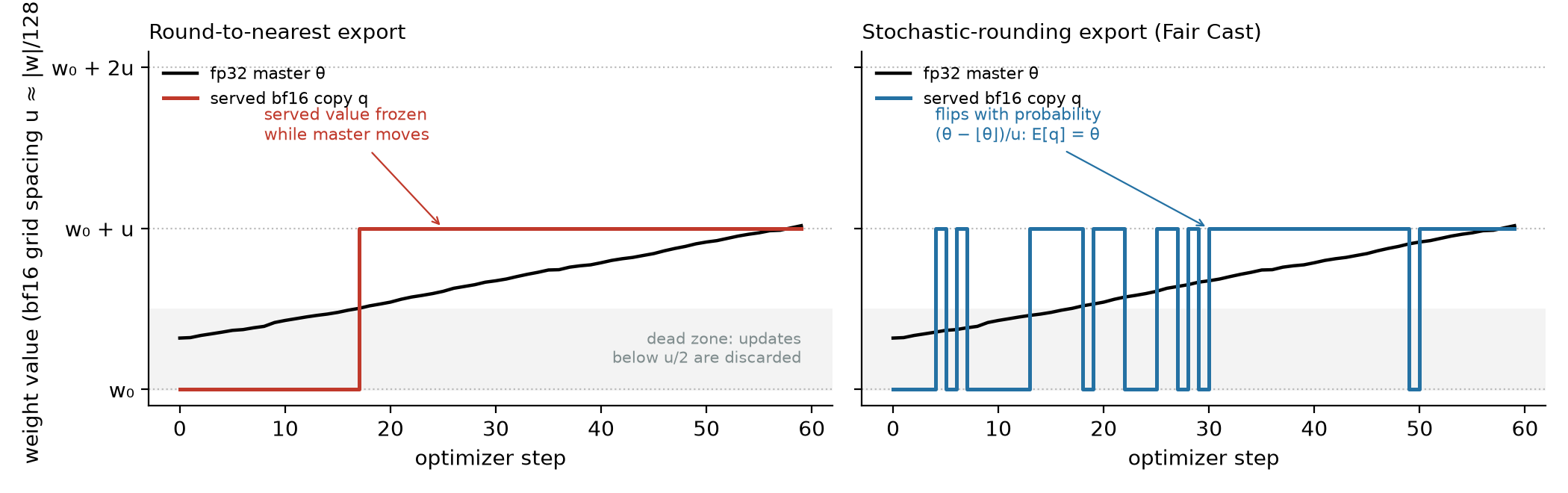}\caption{A weight on the bf16 grid. Left: under round-to-nearest the served value stays at w₀ while the fp32 master accumulates sub-u/2 updates; it jumps only when the master crosses a midpoint. Right: under stochastic rounding the served value flips between neighbours with probability equal to the fractional position, so $E[q] = \theta$ at every step.}\label{fig:bf16grid}\end{figure}

The problem is one of scale. Near a weight of magnitude \textbar w\textbar{} the bf16 grid spacing
is u ≈ \textbar w\textbar/128, so round-to-nearest discards any update smaller than u/2 ≈
\textbar w\textbar/256 (Figure~\ref{fig:bf16grid}). At the learning rates in common use (η ≈ \(10^{-6}\)) almost
every Adam update is that small: on Qwen3-1.7B only 0.4--0.5\% of served weights change per
optimizer step, and the sampler acts on a policy that reflects the optimizer state of 50--150 steps
earlier. Dirhoussi, Gallouédec et al.~\cite{r1} derived the same threshold and observed that the
inference engine ``sees a nearly static model \ldots{} even as the optimizer accumulates meaningful
updates in FP32''; PULSE \cite{r2} measured the resulting sparsity. Both treated it as a
communication artefact and, where they intervened, made the trainer agree with the stale sampler;
the same team now ships only the changed 1\% of bf16 elements as a sparse delta, describing the
sparsity as obtained ``with zero approximation'' \cite{r25}.

Our central observation is that the quantity those diagnostics check is not the quantity that
governs learning. Under round-to-nearest the samples come from the export q and the gradient is
evaluated at q, but the update lands on the master θ, where the optimizer state lives. The policy
that is sampled and differentiated therefore lags the policy the optimizer believes it is improving
by however many updates the dead zone has withheld, and the size of that lag is the ratio
\(\pi_\theta(y)\)/\(\pi_q(y)\). No framework forms it. What frameworks compute is
\(\pi_{\mathrm{trainer}}(y)\)/\(\pi_{\mathrm{sampler}}(y)\), and because the trainer's bf16 forward
pass is itself a function of q, that ratio stays near one however far θ has moved. We measure the
visible ratio at \textbar log ρ\textbar{} ≈ 0.01 per token throughout training while the hidden one
grows to 0.10--0.45 (Figure~\ref{fig:problem}); the per-token effective sample size between master and served policy
falls to 0.2--0.35 on 2,000-token tasks.

\textbf{The fix is to make the export unbiased and to train on it.} Fair Cast replaces
round-to-nearest at the export with stochastic rounding (or error feedback), so that
\(E[q] = \theta\) and no update is systematically withheld, and computes the loss on the exact
served copy, so that the policy generating the data is the policy being differentiated. Figure~\ref{fig:grid}
summarizes the effect across the tasks, model sizes and optimizers we study.

\textbf{Contributions.}

\begin{enumerate}
\def\labelenumi{\arabic{enumi}.}
\tightlist
\item
  \emph{A hidden lag.} We measure the master-versus-served mismatch that current frameworks never
  form, and show why PPO clipping, truncated importance sampling, IcePop masking and bf16-autocast
  training cannot detect it (§3--§4).
\item
  \emph{Fair Cast.} An unbiased export combined with training on the served copy: one line at the
  export call, no hyperparameters, 22--38 ms per export of 1.7B parameters. Both halves are
  necessary; either alone fails (§5, §7.1).
\item
  \emph{Faster and more stable learning.} Over the 25 matched seed-pairs reported in Table 2 and
  §7.5 (Qwen3-0.6B to 4B, AdamW and Muon), Fair Cast is ahead by more than one point on 18, within
  one point on 6 and behind on 1, with a mean gain of 6.1 points. On zebra puzzles it reaches 0.57 ±
  0.16 over four runs where round-to-nearest collapses to 0.33 ± 0.10; under Muon it leads at every
  checkpoint (§7.2--§7.3).
\item
  \emph{The checkpoint is not the model.} Under round-to-nearest the saved fp32 master can score
  12--50 points below its own served export, and any export finer than bf16 (fp16, FP8 via fp16)
  inherits the difference; under Fair Cast master and export agree (§7.4).
\item
  \emph{Lag as a tunable regularizer.} The lag of round-to-nearest is an uncontrolled low-pass
  filter on the served policy. Restoring it explicitly, as an exponential moving average exported
  with the same unbiased rounding, gives a single knob that recovers and surpasses round-to-nearest
  on the one task where the filter helps (§7.5).
\end{enumerate}

\section{Background and Notation}\label{background-and-notation}

Let \(\theta_t\) ∈ \(\mathbb{R}^d\) denote the fp32 master parameters after t optimizer steps and
\(q_t\) = R(\(\theta_t\)) the bf16 copy held by the sampler, where R is the export rounding rule.
For a prompt x and response y, the policy-gradient objective is J(θ) =
\(E_{y\sim \pi_\theta(\cdot |x)}\){[}r(x, y){]}, with gradient

\begin{equation}
\nabla J(\theta ) = E_{y\sim \pi_{\theta }}[ \nabla \log \pi_{\theta }(y) \cdot A(y) ]
\end{equation}

where A is an advantage estimate (in GRPO \cite{r16}, the within-group normalized reward). If
samples are instead drawn from a behaviour policy μ, an unbiased estimate requires the importance
weight

\begin{equation}
\rho(y) = \pi_{\theta }(y) / \mu(y) = \prod_{k} \pi_{\theta }(y_{k} | y_{<k}) / \mu(y_{k} | y_{<k})
\end{equation}

and PPO-style objectives \cite{r21} clip a per-token version of this ratio to {[}1−ε, 1+ε{]}. In LLM
reinforcement learning, μ = \(\pi_{q_t}\) is the policy of the served copy, and the ratio actually
formed inside the loss is

\begin{equation}
\rho_{\mathrm{loss}}(y) = \pi_{\mathrm{trainer}}(y) / \pi_{\mathrm{sampler}}(y)
\end{equation}

where \(\pi_{\mathrm{trainer}}\) is evaluated by the training framework's forward pass and
\(\pi_{\mathrm{sampler}}\) is the log-probability recorded by the inference engine. Truncated
importance sampling (TIS) \cite{r6} and IcePop \cite{r7} reweight or mask tokens according to (3) in
order to correct kernel-level numerical discrepancies between the two engines. When the trainer's
forward pass is executed in bf16 (whether by autocast \cite{r1} or by evaluating the served copy)
both numerator and denominator of (3) are functions of \(q_t\) alone. \(\rho_{\mathrm{loss}}\) is
then ≈1 regardless of the distance between \(\theta_t\) and \(q_t\), and the quantity relevant to
(1),

\begin{equation}
\rho(y) = \pi_{\theta_{t}}(y) / \pi_{q_{t}}(y)
\end{equation}

is never formed. We refer to (3) as the \emph{visible} ratio and to (4) as the \emph{hidden} ratio.

\section{Round-to-Nearest Export Discards the
Update}\label{round-to-nearest-export-discards-the-update}

\textbf{Dead-zone analysis.} bf16 has a 7-bit significand, so the spacing between representable
values near w is u(w) = \(2^{\lfloor \log_{2}|w|\rfloor - 7}\) ≈ \textbar w\textbar/128.
Round-to-nearest maps an update Δ to zero whenever \textbar Δ\textbar{} \textless{} u/2 ≈
\textbar w\textbar/256. Because the fp32 master lies exactly on the bf16 grid at initialization,
this condition is operative from the first step. The per-step fraction of served weights that change
is \(\Pr[\eta |\hat{m}/\sqrt{\hat{v}}| > |w|/256]\); under the measured distribution of
\(|\hat{m}/\sqrt{\hat{v}}|\) this evaluates to 0.3--0.7\% at η = \(10^{-6}\) for Qwen3-1.7B, and to
0.13\% at η = 3×\(10^{-7}\), scaling approximately as η/\textbar w\textbar. A weight drifting at
mean speed \(|\bar{\Delta }|\) crosses a rounding boundary every τ ≈ \(u/(2|\bar{\Delta }|\)) steps,
which is 50--200 steps at these values; under a decaying learning-rate schedule τ → ∞ and the
sampler freezes while the optimizer continues to move.

\begin{figure}[htbp]\centering\includegraphics[width=\linewidth]{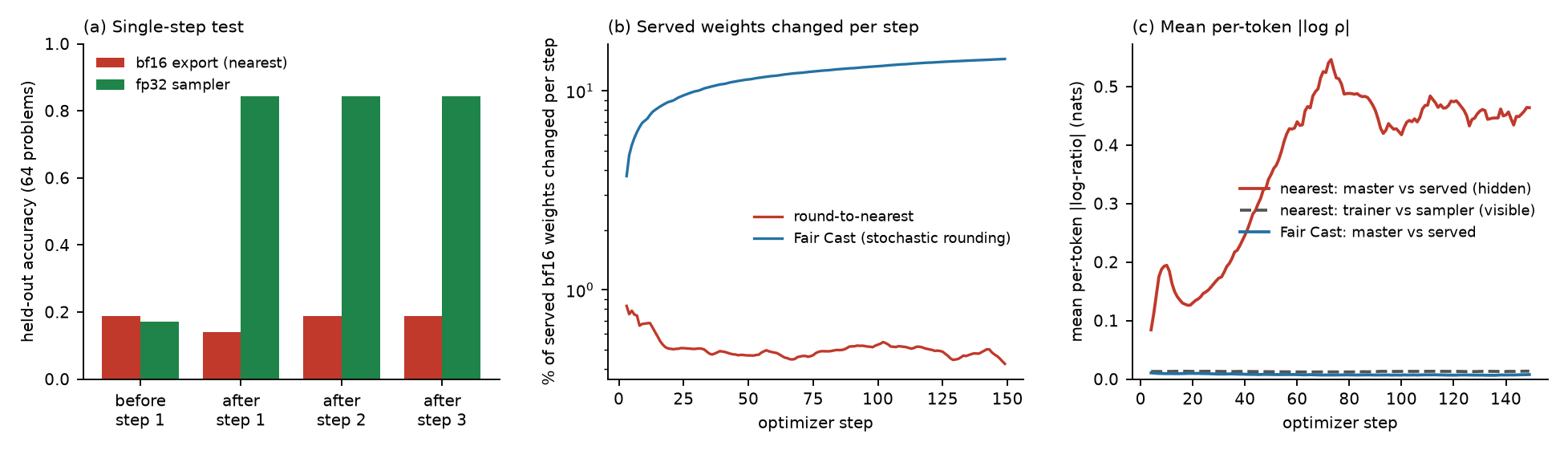}\caption{One optimizer step moves the fp32 policy; the round-to-nearest export does not move, and the mismatch frameworks check does not see it. (a) Matched single-step experiment on knights-and-knaves. (b) Fraction of served bf16 weights that change per optimizer step. (c) Mean per-token |log ρ| on MATH: the visible trainer-versus-sampler ratio stays near 0.01 while the hidden master-versus-served ratio grows to 0.10–0.45 under round-to-nearest and stays at the noise floor under Fair Cast.}\label{fig:problem}\end{figure}

\textbf{Measurements.} The measured per-step fraction of changed served weights on Qwen3-1.7B at η =
\(10^{-6}\) is 0.4--0.5\%. Muon's per-element updates are nearly uniform in magnitude, so at η =
\(10^{-6}\) almost none cross a boundary. On knights-and-knaves, round-to-nearest left the sampler
effectively frozen for the first ≈20 steps (held-out accuracy 0.12 → 0.15 → 0.22 at steps 0/9/19).
Fair Cast moved immediately (0.12 → 0.51 → 0.79). A controlled experiment isolates the mechanism
(Figure~\ref{fig:problem}a). On the same 64 knights-and-knaves problems, a single optimizer step at η = \(10^{-6}\)
moved an fp32 sampler from 0.17 to 0.84. The round-to-nearest bf16 sampler remained at 0.19 across
three successive steps. Figure~\ref{fig:problem}b shows the per-step fraction of served weights that change under
the two export rules over a full run.

\section{The Hidden Lag Between Master and Served
Policy}\label{the-hidden-lag-between-master-and-served-policy}

We log both ratios (3) and (4) throughout training (Figure~\ref{fig:problem}c). On Qwen3-1.7B at η = \(10^{-6}\),
the mean per-token \textbar log \(\rho_{\mathrm{loss}}\)\textbar{} remains ≈0.01 for the entire run,
while the mean per-token \textbar log ρ\textbar{} between master and sampler grows from ≈0 to
0.10--0.16 on MATH and to ≈0.45 on knights-and-knaves; under Fair Cast the same quantity stays
≈0.01. The per-token effective sample size,

\begin{equation}
\mathrm{ESS} = ( \sum_{i} \rho(y_{i}) )^{2} / \sum_{i} \rho(y_{i})^{2}
\end{equation}

computed over the response tokens of a batch with ρ the per-token ratio (4) and normalized by the
token count, is 0.2--0.35 on tasks with 2,000-token responses and ≈0.9 on tasks with 500-token
responses: the per-token gap is larger where the policy has moved further. We use it as a diagnostic
of how far the served policy has fallen behind the optimizer, not as an importance-sampling
correction; the gradient at q with samples from q needs none. Lower learning rates reduce the
visible ratio (as reported in \cite{r8}) while increasing the hidden one, because fewer updates
clear the dead zone. The two observations are consistent once the distinction between (3) and (4) is
made.

\section{Fair Cast}\label{fair-cast}

Fair Cast replaces the export rounding rule R with an unbiased rule and evaluates the training loss
on the served copy. Both components are required.

\textbf{Stochastic rounding.} For θ lying between adjacent bf16 values \(\lfloor \theta \rfloor\)
and \(\lfloor \theta \rfloor\) + u, export the upper value with probability p = (θ −
\(\lfloor \theta \rfloor\))/u. Then \(E[q(\theta )] = \theta\) exactly, and an update of magnitude Δ
changes the export probability by Δ/u, so the expected movement of the served weight equals the true
movement at any learning rate. The implementation adds 16 uniformly distributed random bits to the
fp32 bit pattern and truncates (\texttt{sr\_cast\_bf16}).

\textbf{Error feedback.} Maintain a residual r per weight: \(q_t\) = round(\(\theta_t\) +
\(r_{t- 1}\)), \(r_t\) = \(\theta_t\) + \(r_{t- 1}\) − \(q_t\), with \textbar{}\(r_t\)\textbar{} ≤
u/2. Since \(q_t\) − \(\theta_t\) = \(r_{t- 1}\) − \(r_t\), the sum telescopes:

\begin{equation}
\sum_{s\le t} (q_{s} - \theta_{s}) = r_{- 1} - r_{t}
\end{equation}

so the accumulated export error stays bounded by u for all time: no update is withheld, and the
served copy is unbiased over time rather than per sync. Error feedback is a first-order sigma-delta
modulator; it is deterministic, but it dithers in the same way, flipping a weight whose master sits
off the grid on alternate syncs (\texttt{ef\_cast\_bf16}; one additional fp32 tensor per parameter).

\textbf{Training on the served copy.} The trainer's forward pass uses the served bf16 copy \(q_t\),
with gradients propagated to \(\theta_t\) through a straight-through estimator. The sampled and
differentiated policies are then the same object, \(q_t\); the master receives the gradient
evaluated at \(q_t\). This component is not optional: error-feedback export without it collapsed to
0.00 held-out accuracy in 4 of 4 long runs, and stochastic-rounding export without it degenerated on
Qwen3-0.6B GSM8K and MATH (§7.1). Under either unbiased rule, the rounding perturbation enters the
visible ratio (3) as noise unless the trainer evaluates the same copy the sampler used. Two
implementations are equivalent in value: substituting \(q_t\) for the parameters during the forward
call, or writing \(q_t\) into the parameter storage for the duration of the step and restoring the
master before the optimizer update. With activation checkpointing the second is the exact one, since
the recomputed backward pass then also sees \(q_t\); the first evaluates the backward pass at the
nearest rounding of the master, within one grid spacing of \(q_t\). The main runs use the first; the
second gives the same learning curves on Qwen3-0.6B (knights-and-knaves 0.977 against 0.967, GSM8K
0.756 against 0.742, seed 0; Appendix A).

\textbf{Cost and theoretical properties.} The change is confined to the weight-export call. Measured
cost per full export of 1.7B parameters on an H100 is 22 ms (stochastic rounding, bit-manipulation
kernel) to 38 ms (error feedback), compared with 4.3 ms for round-to-nearest and 31--130 s per
optimizer step (§7.7). Error feedback retains one fp32 residual per parameter (6.9 GB at 1.7B). No
hyperparameters are introduced and the optimizer, loss, sampler kernels, and framework are
unchanged. Per sync, stochastic rounding and error feedback perturb a weight by up to u
(round-to-nearest by at most u/2), with variance at most \(u^{2}\)/4; what distinguishes them from
round-to-nearest is not the size of the perturbation but that it is zero-mean per sync (stochastic
rounding) or zero-mean over time (error feedback) instead of correlated across steps. Fair Cast is
inert whenever the learning rate is large enough that updates clear the dead zone unaided.

\textbf{What is and is not guaranteed.} Unbiasedness is a statement about weights, not about the
policy: \(E[q] = \theta\) does not imply E{[}\(\pi_q\){]} = \(\pi_\theta\), because the network is
nonlinear. The precise statements are the following. (i) The rollouts are exactly on-policy for
\(q_t\), the policy whose log-probabilities enter the loss, so the policy-gradient estimator at
\(q_t\) carries no importance-weighting error; this is the role of training on the served copy. (ii)
The difference between \(\pi_q\) and \(\pi_\theta\) is a rounding perturbation of each weight by
less than one grid spacing u ≈ \textbar w\textbar/128. Under round-to-nearest this perturbation is
sign-coherent across steps, so its first-order effect ∇π·(q − θ) accumulates for the 50--150 steps
of the lag. Under stochastic rounding it is zero-mean and independent across steps, so the
first-order term vanishes in expectation and the remaining effect is the second-order Jensen term
\(\tfrac{1}{2} \operatorname{tr}(H \Sigma )\), with \(\Sigma \preceq (u^{2}/4) I\). (iii) The
gradient passed to θ through the straight-through estimator is the gradient at \(q_t\) rather than
at \(\theta_t\), the standard approximation of quantization-aware training, whose bias is bounded by
the same rounding width \cite{r23,r24}. The residual of (ii) and (iii) is measurable: the mean
per-token \textbar log(\(\pi_\theta\)/\(\pi_q\))\textbar{} under Fair Cast remains ≈0.01 throughout
training, the same as the trainer--sampler ratio, against 0.10--0.45 under round-to-nearest (Figure
4c). Finally, the objective being optimized is the expected reward of the exported model, J(θ) =
\(E_q\){[}J(q){]}, which is the model that samples during training and the model that is deployed;
round-to-nearest instead optimizes a master policy that is never served (§7.4).

\textbf{Variants.} \emph{Dithered error feedback} applies stochastic rounding to the error-fed value
while carrying the residual. \emph{EMA export} exports an unbiasedly rounded exponential moving
average of the master with coefficient α, introducing an explicit and tunable lag (§7.5).

\section{Experimental Setup}\label{experimental-setup}

\textbf{Models and optimizers.} Qwen3-0.6B, Qwen3-1.7B (default), Qwen3-4B, OLMoE-1B-7B, and
Qwen2.5-1.5B-Instruct. AdamW (η = \(10^{-6}\), weight decay 0.01) is the default optimizer; Muon
\cite{r19} (η = \(10^{-6}\)) is used where indicated.

\textbf{Objective and schedule.} GRPO-style clipped policy gradient with group size 8, 24 prompts
per step, and 150 optimizer steps (500 for the long runs). Each arm occupies one H100 (vLLM sampler
with sleep mode and a Hugging Face trainer co-located on the same device, lag 1). The rollout budget
is 2,048 tokens (6,144/4,096 for the synthetic games).

\textbf{Tasks.} Reasoning Gym \cite{r17} knights-and-knaves, zebra puzzles, list functions, string
manipulation, and Countdown; MBPP+ and HumanEval+ \cite{r18}; MATH (levels 3--5 or level 5); GSM8K;
and synthetic games with exact solvers (Nim, subtraction games, Kuhn poker decisions). Appendix B
shows one held-out prompt from each of the zebra, knights-and-knaves and Nim tasks.

\textbf{Evaluation.} Greedy decoding on a held-out set every 10 steps: 512 problems where available
(HumanEval+ 164; 128 for the Qwen3-1.7B MATH and Countdown runs). MBPP+ shares 154 of its 378 task
ids with the MBPP split used for training, so MBPP+ accuracy is reported on the remaining 224
problems (121 of the first 128 for the seed-0 pair). All reported curves and endpoints refer to the
served (sampler) copy unless stated otherwise.

\textbf{Seeds.} Where a configuration is run with several seeds, the seed sets the trainer RNG and
the rollout sampler; the training-prompt sequence is generated from a fixed seed and is therefore
the same for every run of a configuration. Multi-seed results thus measure variability due to
sampling and optimization, not due to data order.

\textbf{Arms.} \emph{Nearest} (round-to-nearest export, the default in current frameworks);
\emph{Fair Cast} with stochastic rounding (default) or error feedback, both with training on the
served copy; an \emph{fp32-sampler oracle} with no rounding anywhere; and controls (nearest with a
truncation penalty, TIS, IcePop mask, FP16 sampler, nearest at tuned learning rate, lag 2/8). One
seed per arm unless stated.

\section{Results}\label{results}

\subsection{Isolating the cause}\label{isolating-the-cause}

Table 1 crosses the export rule with the location of the trainer's forward pass. The two aligned
configurations are stable; the two crossed configurations are not. Unbiased export without training
on the served copy is harmful. The error-feedback variant collapsed to 0.00 in 4 of 4 long runs. The
stochastic-rounding variant on Qwen3-0.6B degenerated into token loops
(\texttt{\textbackslash{}boxed\{\textbackslash{}boxed\{…}) on GSM8K (0.61 → 0.13 by step 19) and
into unterminated reasoning on MATH (→ 0.002). In both, the clip fraction rose in step with the
fraction of served weights that flipped at each export, and the gradient norm fell to zero once
every GRPO group became uniformly incorrect. The same configurations with training on the served
copy are healthy (§7.3). Round-to-nearest with training on the served copy is indistinguishable from
round-to-nearest alone, as expected: both copies are then downstream of the same rounding.

\begin{longtable}[]{@{}
  >{\raggedright\arraybackslash}p{(\linewidth - 4\tabcolsep) * \real{0.3333}}
  >{\raggedright\arraybackslash}p{(\linewidth - 4\tabcolsep) * \real{0.3333}}
  >{\raggedright\arraybackslash}p{(\linewidth - 4\tabcolsep) * \real{0.3333}}@{}}
\caption{Both components are necessary: unbiased export without training on the served copy
collapses, and training on the served copy without unbiased export changes nothing. Held-out
accuracy of the served copy by export rule and trainer forward location.}\tabularnewline
\toprule\noalign{}
\begin{minipage}[b]{\linewidth}\raggedright
\end{minipage} & \begin{minipage}[b]{\linewidth}\raggedright
Trainer forward on fp32 master
\end{minipage} & \begin{minipage}[b]{\linewidth}\raggedright
Trainer forward on served copy
\end{minipage} \\
\midrule\noalign{}
\endfirsthead
\toprule\noalign{}
\begin{minipage}[b]{\linewidth}\raggedright
\end{minipage} & \begin{minipage}[b]{\linewidth}\raggedright
Trainer forward on fp32 master
\end{minipage} & \begin{minipage}[b]{\linewidth}\raggedright
Trainer forward on served copy
\end{minipage} \\
\midrule\noalign{}
\endhead
\bottomrule\noalign{}
\endlastfoot
Round-to-nearest export & Current default; hidden lag (§4) & Same policy as the default: the bf16
forward of the master is the nearest export (0.6B: knights-and-knaves 0.889 against 0.865, GSM8K
0.705 against 0.742) \\
Unbiased export (SR or EF) & Collapse: EF 0.00 (4/4 runs); SR 0.6B GSM8K 0.13, MATH 0.002 &
\textbf{Fair Cast} (§7.2--§7.3) \\
\end{longtable}

\subsection{Long-rollout tasks}\label{long-rollout-tasks}

\begin{figure}[htbp]\centering\includegraphics[width=\linewidth]{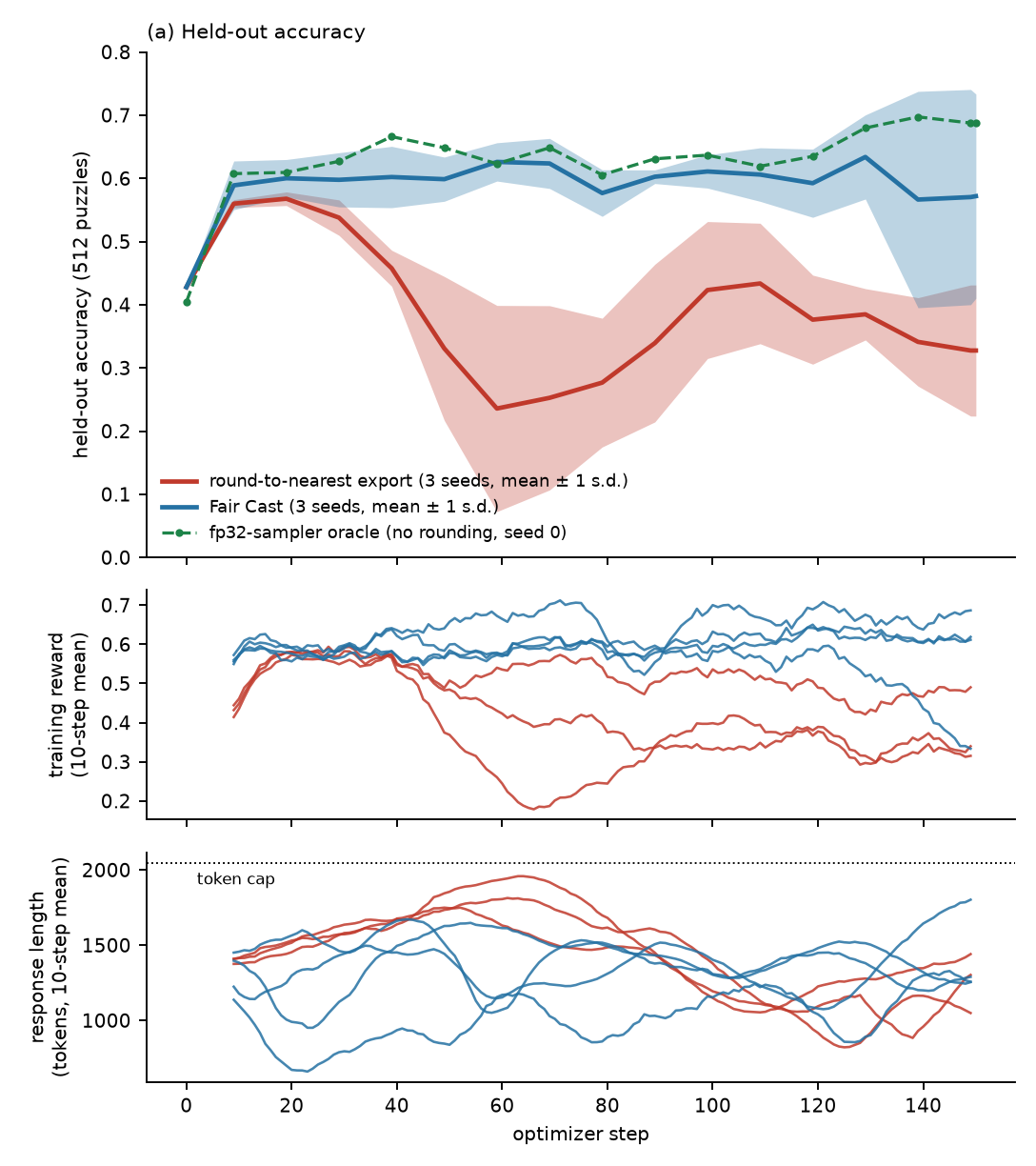}\caption{Fair Cast tracks the fp32-sampler oracle while round-to-nearest lengthens into the token cap and collapses. Zebra puzzles, Qwen3-1.7B, 2,048-token rollouts; three round-to-nearest runs and four Fair Cast runs (seed 0 twice). Top: held-out accuracy (mean ± 1 s.d.). Middle: training reward. Bottom: response length; the dashed line is the token cap.}\label{fig:zebra}\end{figure}

\textbf{Zebra puzzles.} This is the task on which the hidden lag does the most damage.
Round-to-nearest learns normally for about twenty steps, then collapses on every seed: from a peak
of 0.57 ± 0.01 to a final 0.33 ± 0.10. Figure~\ref{fig:zebra} shows why. Between steps 30 and 60 its responses
lengthen into the 2,048-token cap (held-out truncation reaches 0.47--0.89) while training reward
falls, the signature of a policy that is being pushed toward longer derivations by rewards computed
on a model it no longer matches. A DAPO-style truncation penalty \cite{r15} on the round-to-nearest
arm does not prevent the collapse (final 0.45), so the cause is not the penalty structure. Three of
four Fair Cast runs never enter this regime: they end at 0.62--0.70 and track the fp32-sampler
oracle (0.69) throughout, which is the behaviour a correct export should have. A fourth run, a
repeat of seed 0, held 0.53--0.63 for 130 steps and then fell to 0.34 in its last fifteen, so the
four-run mean is 0.57 ± 0.16 against 0.33 ± 0.10 for round-to-nearest.

\textbf{Knights-and-knaves.} Here the lag costs speed rather than stability. With AdamW at 1.7B,
Fair Cast reaches 0.88 at step 9 and ≈0.97 by step 40; round-to-nearest needs about 40 more steps to
get there and ends 3.7 points lower. The gap widens where the dead zone is deepest. With Muon, whose
per-element updates are nearly uniform in magnitude and almost all below u/2, round-to-nearest
leaves the sampler nearly frozen for the first 20 steps and Fair Cast leads at every one of the 15
checkpoints on both seeds (0.988 ± 0.003 versus 0.907 ± 0.016). At 0.6B, where weights are larger
relative to the updates, the gain is 10.2 points; at 4B both exports reach the ceiling (0.992 versus
0.988).

\textbf{Nim.} On a game with an exact solver and a 6,144-token cap, both exports roughly double
their response length and gain about 38 points from step 0, but Fair Cast gets there first: it leads
by 10--30 points through step 100 and peaks at 0.736 against 0.596 before the two converge (0.580
versus 0.570 at step 150). The lag here is a delay, not a different destination.

\subsection{Generalization across model size and
benchmark}\label{generalization-across-model-size-and-benchmark}

\begin{figure}[htbp]\centering\includegraphics[width=\linewidth]{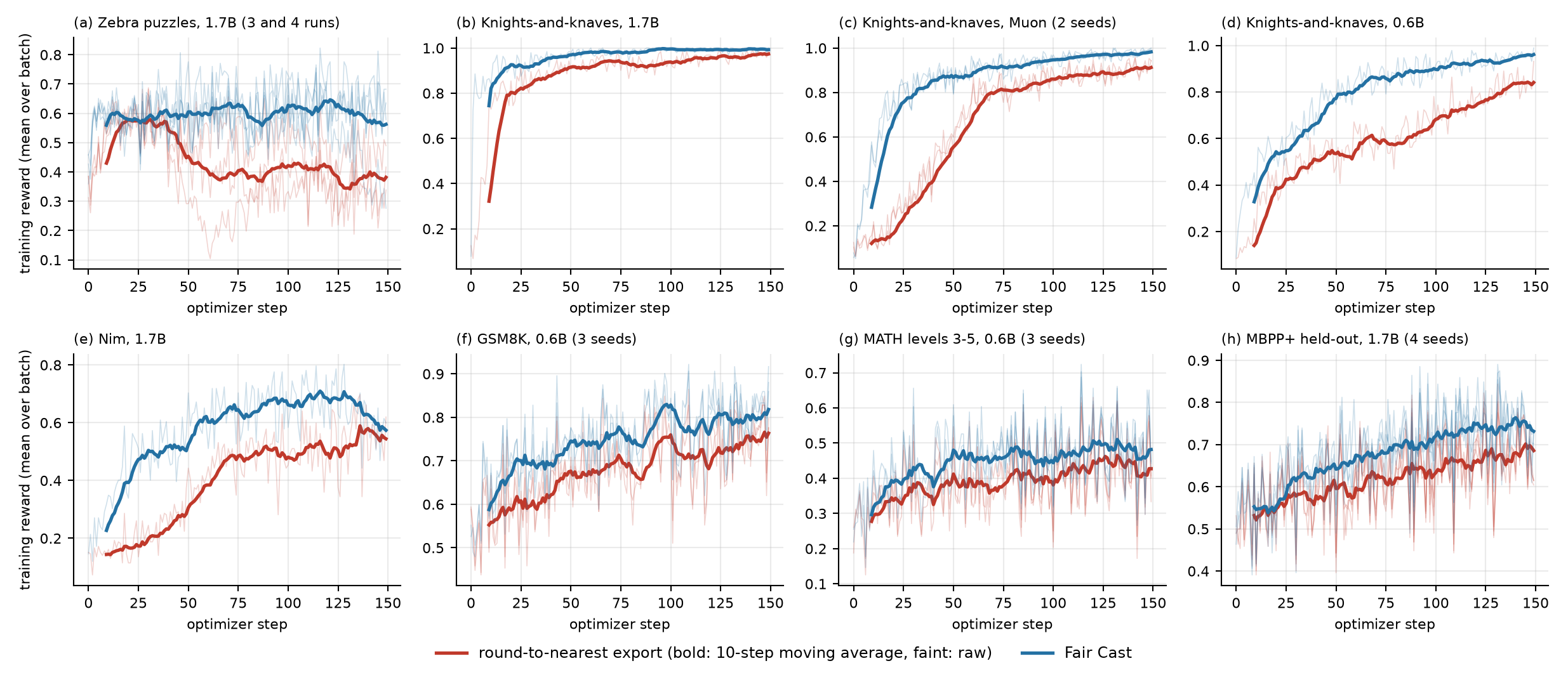}\caption{Training reward tells the same story as held-out accuracy. Mean reward of the training batch against optimizer step for the configurations of Figure~\ref{fig:grid}; faint lines are raw per-step values, bold lines a 10-step moving average (mean across seeds where there are several).}\label{fig:reward}\end{figure}

\begin{figure}[htbp]\centering\includegraphics[width=\linewidth]{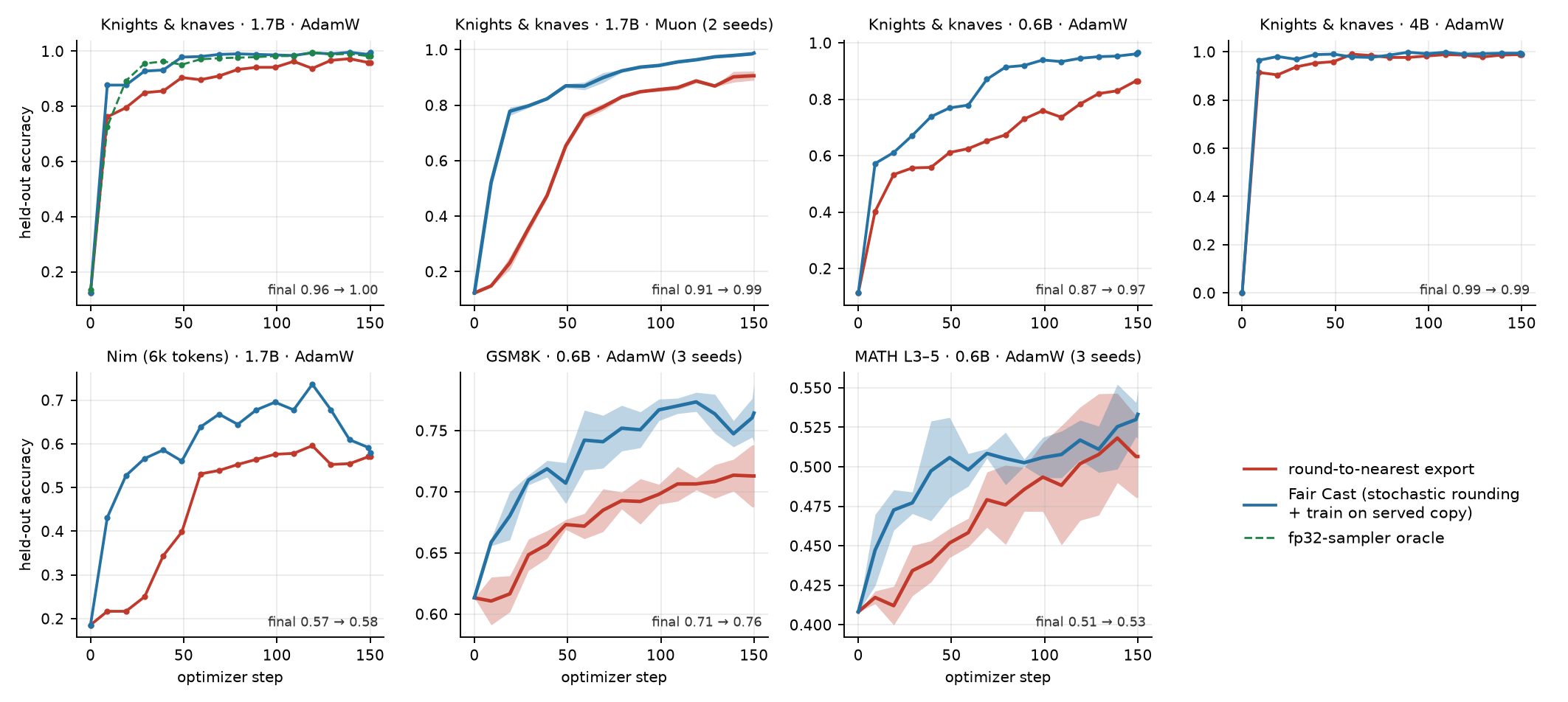}\caption{The gain holds across model sizes, optimizers and task families. Held-out accuracy of the served copy against optimizer step; bands are ± 1 s.d. across seeds; the dashed line is the fp32-sampler oracle where available.}\label{fig:breadth}\end{figure}

Figure~\ref{fig:breadth} shows held-out accuracy against optimizer step for seven configurations spanning model size
(0.6B--4B), optimizer (AdamW, Muon) and task family (logic, games, mathematics); Table 2 gives the
endpoints. Training reward (Figure~\ref{fig:reward}) tells the same story. The pattern is consistent: Fair Cast is ahead at nearly every checkpoint, by the most on
long-rollout logic tasks and by a few points on the recognized benchmarks. On Qwen3-0.6B (three
seeds each) it gains 2.7 points on MATH levels 3--5 and 5.1 on GSM8K, and the curves separate early,
in the first 30 steps, which is where the dead zone withholds the largest fraction of the
optimizer's motion.

\begin{longtable}[]{@{}llllll@{}}
\caption{Fair Cast matches or exceeds round-to-nearest on every task family, with the largest gains
on long-rollout logic tasks. Final held-out accuracy of the served copy after 150 optimizer steps;
mean ± standard deviation when n \textgreater{} 1 seeds; Δ in accuracy points. Qwen3-1.7B with AdamW
unless noted. Zebra: three round-to-nearest runs and four Fair Cast runs (seed 0 run twice). MBPP+:
scored on the 224 problems absent from the MBPP training split (§6). MATH levels 3--5 at 1.7B: a
300-step cosine-decay run reported at its final step; at step 150 it stood at 0.727 against 0.672
(+5.5).}\tabularnewline
\toprule\noalign{}
Task & Model & n & Nearest & Fair Cast & Δ \\
\midrule\noalign{}
\endfirsthead
\toprule\noalign{}
Task & Model & n & Nearest & Fair Cast & Δ \\
\midrule\noalign{}
\endhead
\bottomrule\noalign{}
\endlastfoot
Zebra puzzles & 1.7B & 3 / 4 & 0.33 ± 0.10 & \textbf{0.57 ± 0.16} & +24 \\
Knights-and-knaves & 0.6B & 1 & 0.865 & \textbf{0.967} & +10.2 \\
Knights-and-knaves & 1.7B & 1 & 0.959 & \textbf{0.996} & +3.7 \\
Knights-and-knaves, Muon & 1.7B & 2 & 0.907 ± 0.016 & \textbf{0.988 ± 0.003} & +8.2 \\
Knights-and-knaves & 4B & 1 & 0.988 & 0.992 & +0.4 \\
Nim & 1.7B & 1 & 0.570 & 0.580 & +1.0 \\
MBPP+ (held-out 224) & 1.7B & 4 & 0.595 ± 0.007 & \textbf{0.618 ± 0.017} & +2.3 \\
HumanEval+ & 1.7B & 2 & 0.613 ± 0.039 & 0.625 ± 0.013 & +1.2 \\
String manipulation & 1.7B & 1 & 0.352 & 0.363 & +1.2 \\
MATH, level 5 & 1.7B & 1 & 0.492 & \textbf{0.523} & +3.1 \\
MATH, levels 3--5 & 1.7B & 1 & 0.664 & 0.656 & −0.8 \\
MATH, levels 3--5 & 0.6B & 3 & 0.506 ± 0.025 & \textbf{0.533 ± 0.015} & +2.7 \\
GSM8K & 0.6B & 3 & 0.713 ± 0.025 & \textbf{0.764 ± 0.024} & +5.1 \\
\end{longtable}

Across the 25 matched seed-pairs in Table 2 and §7.5, Fair Cast ends ahead by more than one point on
18, within one point on 6, and behind on 1 (list functions, which §7.5 addresses), with a mean
endpoint difference of +6.1 points. The gain is concentrated on the long-rollout tasks; on
short-answer tasks and tasks near ceiling the two export rules learn the same policy at the same
speed.

\subsection{The checkpoint is not the model}\label{the-checkpoint-is-not-the-model}

\begin{figure}[htbp]\centering\includegraphics[width=\linewidth]{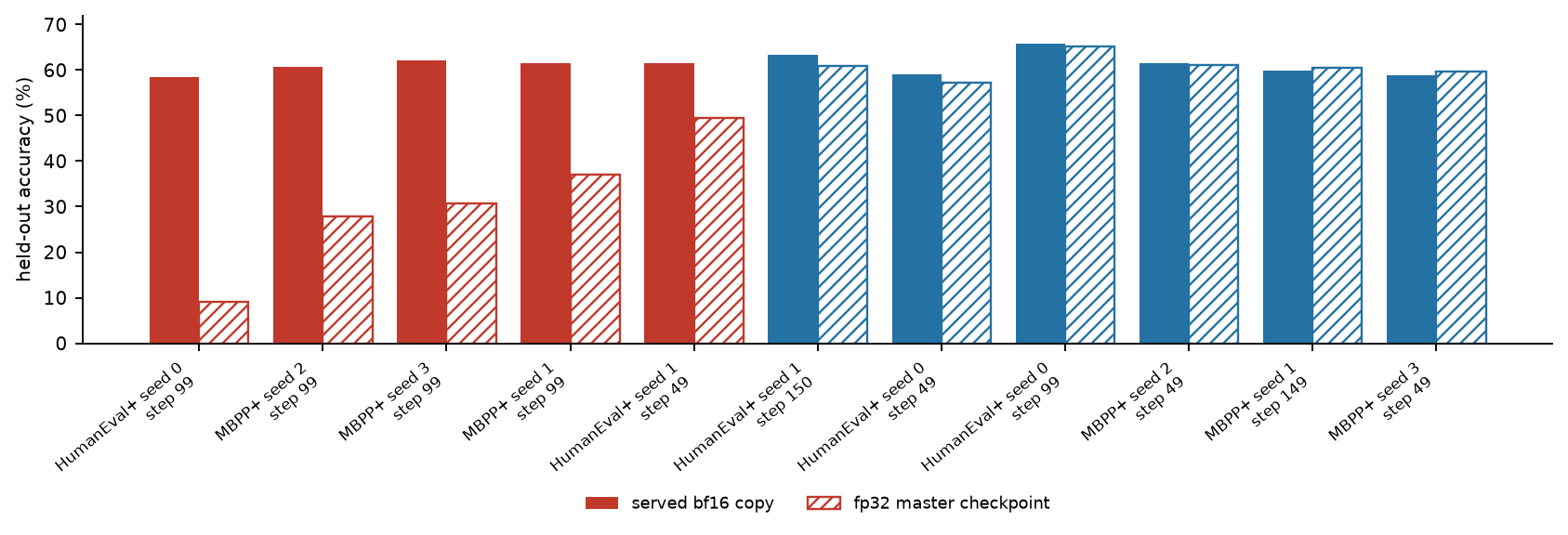}\caption{The fp32 checkpoint is not the served model under round-to-nearest. For each run with master evaluations, held-out accuracy of the served copy and of the fp32 master at the checkpoint of largest gap. Under round-to-nearest the master scores up to 50 points below its own export; under Fair Cast the two agree within 3 points.}\label{fig:ckpt}\end{figure}

A consequence of the lag that has nothing to do with learning curves is that the saved fp32
checkpoint and the model that was actually trained are different objects. At each master-evaluation
checkpoint we score the fp32 master and its served export on the same held-out set (Figure~\ref{fig:ckpt}). Under
round-to-nearest the two agree at step 0 and then drift apart, by up to 50 points on HumanEval+
(served 59.8, master 9.1 at step 99) and 20--30 points on MBPP+. The gap is not monotone: the
sub-threshold residual that the sampler never saw is flushed into the master only when weights cross
bf16 boundaries, so the master can be 30 points below its export at one checkpoint and level with it
50 steps later. Across the 9 round-to-nearest runs with master evaluations, 5 show a gap of at least
12 points at some checkpoint and 2 at the final one. Under Fair Cast the master and its export agree
within 3 points at every checkpoint across 12 runs, because nothing is withheld from the export in
the first place.

\textbf{Which exports inherit the drift.} The practical question is which deployment formats see the
untested residual. We exported saved fp32 masters four ways and evaluated each with greedy decoding
(Table 3). Any format at or coarser than bf16 discards the same residual the sampler never saw and
reproduces the training-time score; any format finer than bf16 (fp16, fp32 or tf32 evaluation,
continued training, or quantization calibrated from fp16) inherits it. A pipeline that trains with
round-to-nearest export and then ships an fp16 model, or resumes training from the fp32 checkpoint,
can deploy a model 30--40 points worse than its training curve indicates, and nothing in the
pipeline signals it. The fp16 export of the HumanEval+ master, for example, produced responses
averaging 1,679 tokens against 344 for the bf16 export: it is a different model.

\begin{longtable}[]{@{}lllll@{}}
\caption{Exports finer than bf16 inherit the drift that the sampler never saw. Held-out accuracy
(\%) of the same saved fp32 master exported four ways; the Fair Cast master gives the same model in
every format.}\tabularnewline
\toprule\noalign{}
Arm & bf16 & fp16 & FP8 via bf16 & FP8 via fp16 \\
\midrule\noalign{}
\endfirsthead
\toprule\noalign{}
Arm & bf16 & fp16 & FP8 via bf16 & FP8 via fp16 \\
\midrule\noalign{}
\endhead
\bottomrule\noalign{}
\endlastfoot
HumanEval+, round-to-nearest & 58.5 & 16.5 & 60.4 & 28.1 \\
Countdown (500 steps), round-to-nearest & 21.9 & 10.2 & 20.3 & 10.2 \\
HumanEval+, Fair Cast & 64.6 & 62.2 & 61.0 & 61.0 \\
\end{longtable}

\subsection{Lag as a tunable regularizer}\label{lag-as-a-tunable-regularizer}

Round-to-nearest export transmits an update only once the accumulated motion of a weight crosses a
bf16 boundary. In signal terms the served policy is a low-pass-filtered version of the master: fast,
small changes are suppressed and only sustained drift passes. This filter has two effects. It is a
bias (the served policy lags, §3--§4), and it is a regularizer: the sampler cannot commit to
whatever direction the reward favours in the current step. Fair Cast removes both. Where the task
rewards early commitment, removing the regularizer is a loss; the filter can be restored explicitly
by exporting an exponential moving average (EMA) of the master, \(m_t\) = (1 − α) \(m_{t- 1}\) + α
\(\theta_t\), with the same unbiased rounding and the same training on the served copy. The
coefficient α sets the lag (≈1/α steps) and replaces an accidental, weight-magnitude-dependent delay
with a chosen one.

List functions (Qwen3-1.7B, 512 held-out problems) is the task on which the filter matters (Figure
8). Fair Cast without lag commits to answer-only outputs by step 29 (median response 13 characters,
every sample a bare \texttt{\textbackslash{}boxed\{…\}}) and plateaus at 0.523; round-to-nearest
keeps producing ≈600-token derivations and reaches 0.668. Fair Cast with the EMA export at α = 0.05
reaches 0.707 and 0.709 on two seeds, above round-to-nearest, with full derivations throughout; α =
0.01 (a longer lag) gives 0.666, matching round-to-nearest. The unbiased export is thus compatible
with as much or as little smoothing as the task needs, and the amount is a hyperparameter rather
than a side effect of the number format.

\begin{figure}[htbp]\centering\includegraphics[width=\linewidth]{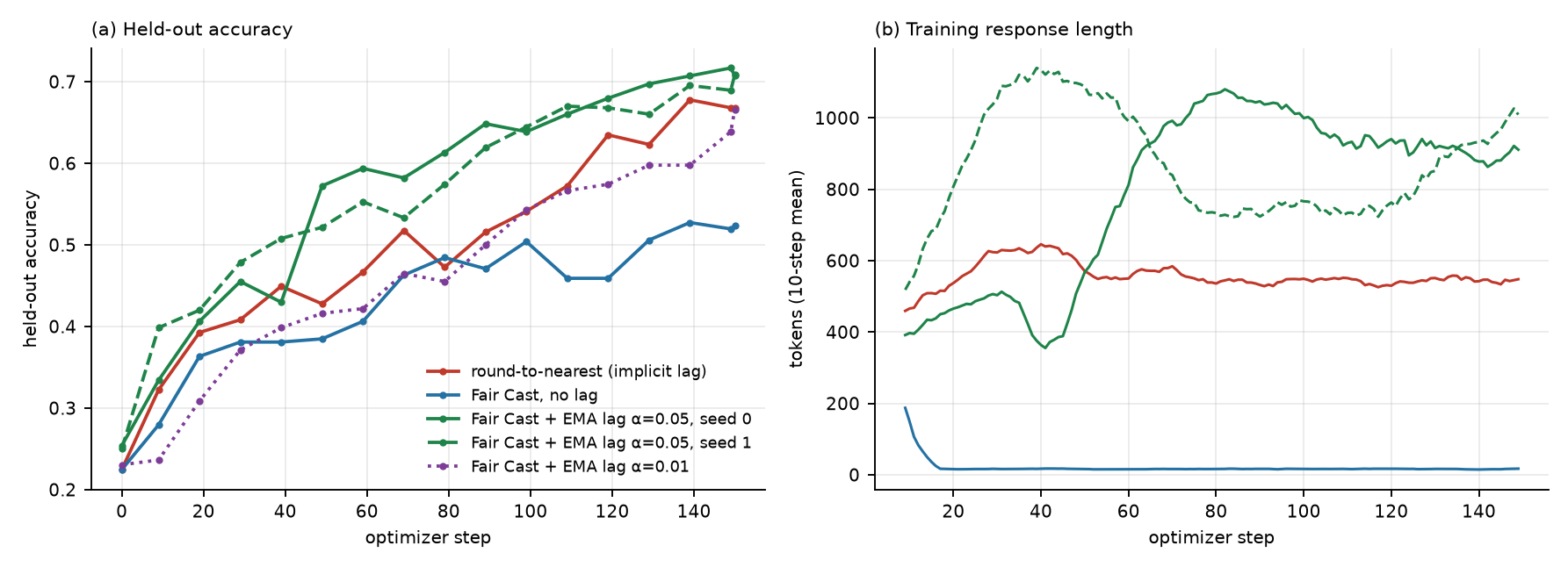}\caption{Lag as a tunable regularizer: list functions, Qwen3-1.7B, 512 held-out problems. Left: held-out accuracy. Right: training response length. Fair Cast without lag commits to answer-only outputs by step 29; with an EMA export at α = 0.05 it keeps full derivations and ends above round-to-nearest on both seeds.}\label{fig:lag}\end{figure}

\subsection{Transfer to AIME24 and AMC23}\label{transfer-to-aime24-and-amc23}

A natural worry is that a policy trained on the served copy learns something specific to the
training distribution. It does not. We evaluated saved 1.7B MATH-trained checkpoints on competition
problems with an 8,192-token budget at temperature 1 (avg@16). On the cosine-schedule pair the Fair
Cast checkpoint transfers better: 0.204 against 0.152 on AIME24, 0.548 against 0.495 on AMC23 and
0.790 against 0.754 on MATH-500, with GSM8K unchanged at 0.83 for both. On the level-5-trained pair
the two are within noise (AIME24 0.163 against 0.165), which with 30 problems and a standard error
of 0.03--0.04 per cell is the expected outcome of an equal pair. Transfer is at least as good as
under round-to-nearest, and neither export rule induces forgetting on GSM8K.

\subsection{Ablations, controls, and cost}\label{ablations-controls-and-cost}

\textbf{Ablations.} The ablations ask whether anything simpler would do. Training on the served copy
is necessary (§7.1). The choice of unbiased rule is not: stochastic rounding and error feedback are
equal on knights-and-knaves and HumanEval+, and error feedback is the stronger of the two only on
the 500-step Countdown run, where its deterministic residual matters most. Fair Cast composes with
TIS, which corrects version lag rather than rounding lag, and the two effects are additive. An FP16
sampler \cite{r3} shrinks the dead zone eightfold but does not remove it and adds overflow risk.
Raising the round-to-nearest learning rate closes part of the speed gap, but at 3 × \(10^{-6}\) and
\(10^{-5}\) it produced two MATH collapses, which is the usual price of pushing a biased sampler
harder. On OLMoE-1B-7B and Qwen2.5-1.5B-Instruct (MATH, 128 problems) the two rules are within
noise, with Fair Cast never worse.

\textbf{Cost.} The export is a negligible fraction of the step. On Qwen3-1.7B (1.72B parameters,
H100, mean of 10 full exports) round-to-nearest takes 4.3 ms, stochastic rounding 21.7 ms with the
bit-manipulation kernel, and error feedback 37.2 ms plus 6.9 GB of fp32 residual; a naive
seven-operation stochastic-rounding kernel takes 136.6 ms. Against an optimizer step of 31--130 s in
our runs this is below 0.1\%. Wall-clock time per step is identical between the two rules wherever
output lengths match (knights-and-knaves 56.2 against 56.0 s), the larger differences on list
functions and Countdown reflect policy length, and FLOPs are identical by construction. The method
is free in the sense that matters: it changes what the sampler sees, not how much work the loop
does.

\section{Related Work}\label{related-work}

\textbf{The export lag.} Dirhoussi, Gallouédec et al.~\cite{r1} derive K ≈ \textbar w\textbar/(256η)
and describe the ``nearly static model''; their remedy, bf16-autocast training, makes the trainer
agree with the stale sampler: the visible ratio becomes ≈1 and the hidden ratio is unchanged. PULSE
\cite{r2}, ``Lost in the bf16 cast'', QaRL, and MXFP8-RL \cite{r5} measure trainer--sampler
numerical mismatch at matched weights; TIS \cite{r6} and IcePop \cite{r7} correct it with importance
weights or masks, all downstream of the same export. Qi et al.~\cite{r3} advocate FP16 to shrink the
mismatch; §7.4 shows that an fp16 export of a round-to-nearest-trained master inherits the master's
drift. Ke et al.~\cite{r4} align rounding locations between training and inference kernels, which
addresses the visible ratio. The finding of \cite{r8} that lower learning rates reduce mismatch
refers to the visible ratio (§4). Delta weight sync in TRL \cite{r25} builds infrastructure on the
same observation, transmitting only the roughly 1\% of bf16 elements that change per step; the
updates that do not cross a boundary are treated as absent rather than as withheld. The same
arithmetic has been observed outside RL: in quantization-aware fine-tuning of ternary models, a bf16
cast at export reverts low-learning-rate code changes and can erase most of the fine-tuning gain,
and stochastic-rounding or compensated writes recover it \cite{r26}; in multi-teacher on-policy
distillation, 97\% of fp32 master weights move while only 7--11\% of their bf16 images do
\cite{r27}. Neither work examines the RL sampler, and \cite{r26} states that exposure of RL
fine-tuning to the effect is a hypothesis; the present paper tests it.

\textbf{Stochastic rounding and error feedback} are established tools in low-precision training
\cite{r9} and in communication-compressed distributed SGD \cite{r10,r11,r12}. We apply them at the
trainer-to-sampler export rather than to gradients or optimizer states. The only prior proposal at
this location of which we are aware is a pair of unmerged prime-rl pull requests \cite{r13} applying
stochastic rounding to an FP8 export ``so the served value equals the BF16 trainer value in
expectation''. They do not address the bf16 case, error feedback, or training on the served copy,
and they report no learning results. QCRL \cite{r20} concerns quantization-consistent RL objectives
and is unrelated. Schulman et al.~\cite{r14} observe that ``RL requires very low capacity \ldots{}
O(1) bits per episode'' and that ``the optimal LR for LoRA is consistently 10× the one used for
FullFT''. Both observations place full-parameter RL post-training deeper in the dead zone, and they
predict that separately served LoRA adapters would largely escape it.

\section{Discussion}\label{discussion}

\textbf{Why a 0.4\% perturbation matters.} Rounding a weight to bf16 perturbs it by at most
\textbar w\textbar/256. In a forward pass that perturbation is negligible, which is why bf16 serving
works. In a training loop it is not negligible, for two reasons. First, round-to-nearest applies it
coherently: the same weights are held back in the same direction for 50--150 steps, so the served
policy is a stale copy rather than a noisy one, and the staleness compounds over 2,000-token
sequences into an effective sample size of 0.2--0.35. Second, the optimizer never learns of the
discrepancy, because the gradient is computed on a copy that shares the rounding. Fair Cast changes
the first fact (the perturbation becomes zero-mean and independent across steps) and the second (the
gradient is computed on the copy that generated the data).

\textbf{Fix the sampler or degrade the trainer?} The remedies proposed in prior work (autocast
training, FP16 throughout, aligned rounding locations) make the trainer agree with the sampler; they
close the visible ratio and leave the hidden one unchanged. Fair Cast makes the sampler an unbiased
image of the trainer and trains the policy that is actually served. The two approaches are
complementary: Fair Cast composes with TIS-style corrections for the kernel-level mismatch that
remains.

\textbf{Trade-off with sparse weight transfer.} Fair Cast changes 2--15\% of served weights per step
instead of 0.4--0.5\%, so a delta-sync scheme that ships only changed elements \cite{r25} would
compress 5--30× less. The two are not in conflict: the delta is still an order of magnitude smaller
than the full model, and the cost buys a sampler that is no longer 50--150 steps stale.

\textbf{What to change in practice.} Export with stochastic rounding and compute the loss on the
served copy; the change is confined to the export call and costs tens of milliseconds per step.
Evaluate the served copy, not the fp32 checkpoint. Deploy from bf16 or from FP8 calibrated on bf16;
do not ship an fp16 model or resume training from the fp32 master after round-to-nearest training
without re-evaluating. If a task benefits from the implicit smoothing of round-to-nearest, restore
it deliberately with the EMA export rather than by accident through the number format.

\textbf{Limitations.} The models studied have at most 4B parameters, the experiments share one
training implementation, and several configurations were run with a single seed; evaluation-set
sizes are given in §6, and the standard error of a single endpoint difference is about 3 points at
512 problems and 6 at 128, so individual differences inside that range are not significant on their
own. We have not characterized in advance which tasks need the explicit lag of §7.5.

\section{Conclusion}\label{conclusion}

Round-to-nearest conversion at the weight export discards most RL updates at the learning rates in
common use, leaving the sampler tens to hundreds of steps behind the optimizer. The diagnostics in
current frameworks cannot observe the resulting mismatch, because they compare two copies downstream
of the same rounding. Replacing the rounding rule with an unbiased one and training on the served
copy makes the rollouts exactly on-policy for the policy being differentiated and removes the
systematic lag of the export, at negligible cost and with no new hyperparameters. In our experiments
this change improves endpoint accuracy on long-rollout tasks, averts a collapse that
round-to-nearest exhibits consistently on zebra puzzles, and removes the disagreement between the
fp32 checkpoint and the served model. It also removes an implicit regularizer, and in two
configurations the removal is harmful unless an explicit lag is reintroduced. Both effects follow
from the same observation: the served copy, not the master, is the policy being trained.

\section*{Reproducibility}\label{reproducibility}
\addcontentsline{toc}{section}{Reproducibility}

Code, configurations, saved fp32 master checkpoints, final served copies, and held-out responses for
every checkpoint will be released with the paper.

\appendix
\setcounter{table}{0}\renewcommand{\thetable}{A\arabic{table}}

\section{Experimental settings}\label{experimental-settings}

\begin{longtable}[]{@{}
  >{\raggedright\arraybackslash}p{(\linewidth - 2\tabcolsep) * \real{0.2500}}
  >{\raggedright\arraybackslash}p{(\linewidth - 2\tabcolsep) * \real{0.7500}}@{}}
\caption{Training and evaluation configuration shared by all runs unless stated in the
text.}\tabularnewline
\toprule\noalign{}
\begin{minipage}[b]{\linewidth}\raggedright
Setting
\end{minipage} & \begin{minipage}[b]{\linewidth}\raggedright
Value
\end{minipage} \\
\midrule\noalign{}
\endfirsthead
\toprule\noalign{}
\begin{minipage}[b]{\linewidth}\raggedright
Setting
\end{minipage} & \begin{minipage}[b]{\linewidth}\raggedright
Value
\end{minipage} \\
\midrule\noalign{}
\endhead
\bottomrule\noalign{}
\endlastfoot
Objective & GRPO-style clipped policy gradient, group size 8, clip 0.2 \\
Prompts per optimizer step & 24 (192 rollouts) \\
Optimizer steps & 150 (500 for the Countdown long run; 300 with cosine decay to \(10^{-7}\) for the
1.7B MATH levels 3--5 pair) \\
Learning rate & \(10^{-6}\) constant, AdamW (β = 0.9, 0.999, weight decay 0.01); Muon at \(10^{-6}\)
with momentum 0.95 and RMS-matched updates (0.2·√max(m, n)) on 2-D hidden weights, AdamW at
\(10^{-6}\) for embeddings, norms and the output head, where indicated \\
Rollout budget & 2,048 tokens (6,144 for Nim); temperature 1.0 \\
Sampler / trainer & vLLM bf16 sampler and Hugging Face trainer on one H100, lag 1 \\
Export & round-to-nearest, stochastic rounding (bit-manipulation kernel), or error feedback, applied
to every parameter after each optimizer step \\
Trainer forward on served copy & parameter substitution for the forward call (main runs); the
parameter-storage variant, which also routes the checkpointed backward pass through \(q_t\),
reproduces it on Qwen3-0.6B (knights-and-knaves 0.977 vs 0.967, GSM8K 0.756 vs 0.742) \\
Held-out evaluation & greedy decoding every 10 steps; 512 problems where available (MBPP+ 224 after
removing training overlap, HumanEval+ 164, 128 for the 1.7B MATH and Countdown runs) \\
Seeds & vary the trainer RNG and the rollout sampler; the training-prompt sequence is fixed \\
\end{longtable}

\section{Task examples}\label{task-examples}

One held-out prompt from each of the three logic and game tasks, exactly as the model receives it
(the chat template is applied around it). Zebra puzzles and knights-and-knaves are drawn from
Reasoning Gym \cite{r17}; the reward is 1 if the boxed answer matches the reference under the task's
own checker, and 0 otherwise. Nim positions are generated with an exact solver; the reward is 1 for
any winning move, or for \texttt{none} when no winning move exists.

\textbf{Zebra puzzle} (reference answer: \texttt{bob}).

\begin{quote}\small\ttfamily\raggedright\setlength{\parskip}{0pt}
This is a logic puzzle. There are 4 houses (numbered 1 on the left, 4 on the right), from the perspective of someone standing across the street from them. Each has a different person in them. They have different characteristics: \\
 - Each person has a unique name: bob, arnold, alice, carol \\
 - Each mother is accompanied by their child: timothy, bella, alice, billy \\
 - Each person has a favorite drink: coffee, milk, boba tea, water \\
 - The people are of nationalities: dane, german, chinese, brit \\
~ \\
1. The person's child is named Alice is the boba tea drinker. \\
2. The coffee drinker is Carol. \\
3. The person who is the mother of Billy is the one who only drinks water. \\
4. Bob is the Dane. \\
5. Carol is directly left of the German. \\
6. The British person is the person's child is named Bella. \\
7. The Chinese is Alice. \\
8. The Chinese is in the fourth house. \\
9. Arnold is the boba tea drinker. \\
10. The one who only drinks water and the person's child is named Bella are next to each other. \\
~ \\
What is Name of the person who lives in House 1?? Provide only the name of the person as your final answer. \\
~ \\
Give the final answer inside \textbackslash{}boxed\{\}.
\end{quote}

\textbf{Knights-and-knaves} (reference answer:
\texttt{Lily\ is\ an\ egoist,\ and\ Logan\ is\ an\ altruist.}).

\begin{quote}\small\ttfamily\raggedright\setlength{\parskip}{0pt}
A very special island is inhabited only by altruists and egoists. Altruists always tell the truth, and egoists always lie. You meet 2 inhabitants: Lily, and Logan. Lily noted, \textquotedbl{}Logan is an egoist\textquotedbl{}. \textquotedbl{}Lily is an altruist or Lily is an egoist,\textquotedbl{} Logan declared. So who is an altruist and who is an egoist? (Format your answer like: \textquotedbl{}Lily is a altruist/egoist, and Logan is a altruist/egoist\textquotedbl{}) \\
~ \\
Give the final answer inside \textbackslash{}boxed\{\}.
\end{quote}

\textbf{Nim} (heaps \cite{r5,r13,r2,r3}; the only winning move is \texttt{2:9}).

\begin{quote}\small\ttfamily\raggedright\setlength{\parskip}{0pt}
Nim (normal play: the player who takes the last stone wins). Heaps: \cite{r5,r13,r2,r3}. It is your move; a move removes 1 or more stones from exactly one heap. If a winning move exists, give one as heap\_index:stones\_removed with 1-based heap index (for example \textbackslash{}boxed\{2:5\}). If no winning move exists, answer \textbackslash{}boxed\{none\}. \\
~ \\
Give the final answer inside \textbackslash{}boxed\{\}.
\end{quote}

\begingroup\small\setlength{\bibsep}{3pt}

\endgroup


\begin{thebibliography}{27}
\bibitem{r1} A. Dirhoussi, Q. Gallouédec, E. Beeching, L. Tunstall, K. Rasul. Defeating the trainer-generator precision mismatch in TRL. Hugging Face, April 2026. \url{https://aminedirohf-trainer-generator-bf16-mismatch.hf.space}
\bibitem{r2} E. Miahi, E. Belilovsky. Understanding and exploiting weight update sparsity for communication-efficient distributed RL (PULSE). \href{https://arxiv.org/abs/2602.03839}{arXiv:2602.03839}, 2026.
\bibitem{r3} P. Qi, Z. Liu, X. Zhou, T. Pang, et al. Defeating the training-inference mismatch via FP16. \href{https://arxiv.org/abs/2510.26788}{arXiv:2510.26788}, 2025.
\bibitem{r4} Y. Ke, T. Cai, R. Wang. Where rounding happens matters: rounding-location alignment stabilizes LLM reinforcement learning. \href{https://arxiv.org/abs/2610.32951}{arXiv:2610.32951}, 2026.
\bibitem{r5} W. Zhang. End-to-end MXFP8 reinforcement learning on Blackwell: what matching buys, when it pays off, and how it fails. \href{https://arxiv.org/abs/2610.34915}{arXiv:2610.34915}, 2026.
\bibitem{r6} F. Yao, et al. Your efficient RL framework secretly brings you off-policy RL training. Technical blog, 2025. \url{https://fengyao.notion.site/off-policy-rl}
\bibitem{r7} MiniMax. IcePop: masking tokens whose training and inference probabilities disagree. MiniMax-M2 technical blog, 2025.
\bibitem{r8} Y. Zhang, Y. Li, J. Liu, J. Xu, et al. Beyond precision: training-inference mismatch is an optimization problem and simple LR scheduling fixes it. \href{https://arxiv.org/abs/2602.01826}{arXiv:2602.01826}, 2026.
\bibitem{r9} S. Gupta, A. Agrawal, K. Gopalakrishnan, P. Narayanan. Deep learning with limited numerical precision. ICML, 2015. \href{https://arxiv.org/abs/1502.02551}{arXiv:1502.02551}.
\bibitem{r10} F. Seide, H. Fu, J. Droppo, G. Li, D. Yu. 1-bit stochastic gradient descent and its application to data-parallel distributed training of speech DNNs. Interspeech, 2014. \url{https://www.isca-archive.org/interspeech\_2014/seide14\_interspeech.html}
\bibitem{r11} S. U. Stich, J.-B. Cordonnier, M. Jaggi. Sparsified SGD with memory. NeurIPS, 2018. \href{https://arxiv.org/abs/1809.07599}{arXiv:1809.07599}.
\bibitem{r12} P. Richtárik, I. Sokolov, I. Fatkhullin. EF21: a new, simpler, theoretically better, and practically faster error feedback. NeurIPS, 2021. \href{https://arxiv.org/abs/2106.05203}{arXiv:2106.05203}.
\bibitem{r13} Prime Intellect. prime-rl pull request \#3628: make online FP8 weight updates safe and unbiased (stochastic rounding at export). 2026. \url{https://github.com/PrimeIntellect-ai/prime-rl/pull/3628}
\bibitem{r14} J. Schulman, et al. LoRA without regret. Thinking Machines Lab, 2025. \url{https://thinkingmachines.ai/blog/lora/}
\bibitem{r15} Q. Yu, Z. Zhang, R. Zhu, Y. Yuan, et al. DAPO: an open-source LLM reinforcement learning system at scale. \href{https://arxiv.org/abs/2503.14476}{arXiv:2503.14476}, 2025.
\bibitem{r16} Z. Shao, P. Wang, Q. Zhu, R. Xu, et al. DeepSeekMath: pushing the limits of mathematical reasoning in open language models. \href{https://arxiv.org/abs/2402.03300}{arXiv:2402.03300}, 2024.
\bibitem{r17} Z. Stojanovski, O. Stanley, J. Sharratt, et al. Reasoning Gym: reasoning environments for reinforcement learning with verifiable rewards. NeurIPS, 2025. \href{https://arxiv.org/abs/2505.24760}{arXiv:2505.24760}.
\bibitem{r18} J. Liu, C. S. Xia, Y. Wang, L. Zhang. Is your code generated by ChatGPT really correct? Rigorous evaluation of large language models for code generation. NeurIPS, 2023. \href{https://arxiv.org/abs/2305.01210}{arXiv:2305.01210}.
\bibitem{r19} K. Jordan, et al. Muon: an optimizer for hidden layers in neural networks. 2024. \url{https://kellerjordan.github.io/posts/muon/}
\bibitem{r20} QCRL: quantization-consistent reinforcement learning for LLM reasoning. OpenReview, 2026 (under review). \url{https://openreview.net/forum?id=pX7KzJPRpY}
\bibitem{r21} J. Schulman, F. Wolski, P. Dhariwal, A. Radford, O. Klimov. Proximal policy optimization algorithms. \href{https://arxiv.org/abs/1707.06347}{arXiv:1707.06347}, 2017.
\bibitem{r22} T. Dettmers, M. Lewis, S. Shleifer, L. Zettlemoyer. 8-bit optimizers via block-wise quantization. ICLR, 2022. \href{https://arxiv.org/abs/2110.02861}{arXiv:2110.02861}.
\bibitem{r23} Y. Bengio, N. Léonard, A. Courville. Estimating or propagating gradients through stochastic neurons for conditional computation. \href{https://arxiv.org/abs/1308.3432}{arXiv:1308.3432}, 2013.
\bibitem{r24} P. Yin, J. Lyu, S. Zhang, S. Osher, Y. Qi, J. Xin. Understanding straight-through estimator in training activation quantized neural nets. ICLR, 2019. \href{https://arxiv.org/abs/1903.05662}{arXiv:1903.05662}.
\bibitem{r25} A. Dirhoussi, Q. Gallouédec, K. Rasul, L. Tunstall, E. Beeching, et al. Shipping a trillion parameters with a Hub bucket: delta weight sync in TRL. Hugging Face, May 2026. \url{https://huggingface.co/blog/delta-weight-sync}
\bibitem{r26} A. Sahai, N. Raj, A. Srivastava. Lost in the bf16 cast: exporting ternary language models can revert most low-learning-rate code changes. \href{https://arxiv.org/abs/2610.07853}{arXiv:2610.07853}, 2026.
\bibitem{r27} S. Zhu, S. Huang, K. Zhang, Y. Yang, et al. From gradients to capabilities: understanding multi-teacher on-policy distillation. \href{https://arxiv.org/abs/2610.02179}{arXiv:2610.02179}, 2026.
\end{thebibliography}
\end{document}